\documentclass[10pt,twocolumn,letterpaper]{article}

\usepackage{wacv}
\usepackage{times}
\usepackage{epsfig}
\usepackage{graphicx}
\usepackage{amsmath}
\usepackage{amssymb}
\usepackage{booktabs}
\usepackage{multirow}
\usepackage{wrapfig}
\usepackage{siunitx}
\usepackage{amssymb}
\usepackage{enumitem}

\wacvfinalcopy 

\def\wacvPaperID{420} 

\ifwacvfinal\fi
\begin{document}

\title{Deep Semantic Instance Segmentation of Tree-like Structures Using Synthetic Data}

\author{Kerry Halupka \\
IBM Research, Level 22/60 City Rd, Southbank, Victoria, Australia.\\
{\tt\small kerry.halupka@au1.ibm.com}
\and
Rahil Garnavi \\
IBM Research, Level 22/60 City Rd, Southbank, Victoria, Australia.\\
{\tt\small rahilgar@au1.ibm.com}
\and
Stephen Moore \\
IBM Research, Level 22/60 City Rd, Southbank, Victoria, Australia.\\
{\tt\small stevemoore@au1.ibm.com}
}

\maketitle
\ifwacvfinal\thispagestyle{empty}\fi

\begin{abstract}
   Tree-like structures, such as blood vessels, often express complexity at very fine scales, requiring high-resolution grids to adequately describe their shape. Such sparse morphology can alternately be represented by locations of centreline points, but learning from this type of data with deep learning is challenging due to it being unordered, and permutation invariant. In this work, we propose a deep neural network that directly consumes unordered points along the centreline of a branching structure, to identify the topology of the represented structure in a single-shot. Key to our approach is the use of a novel multi-task loss function, enabling instance segmentation of arbitrarily complex branching structures. We train the network solely using synthetically generated data, utilizing domain randomization to facilitate the transfer to real 2D and 3D data. Results show that our network can reliably extract meaningful information about branch locations, bifurcations and endpoints, and sets a new benchmark for semantic instance segmentation in branching structures.
\end{abstract}

\section{Introduction}

\label{sec:intro}
In this work, we are interested in the problem of extracting topological information from thin, branching structures, such as blood vessels, neurons, and trees in nature. The complexity of these structures is often expressed at a very fine scale, meaning that a high resolution, sparsely filled grid would be required to adequately describe their shape with pixels or voxels, making analysis computationally expensive. A possible solution to this issue is to represent the structures as centreline points in Euclidean Space, which would convert the large, sparse grid to a manageable form that allows more fine-grained structure to be represented. However, dealing with a set of points that are unordered and permutation invariant introduces some challenges. In particular, traditional deep learning architectures for geometric 3D data normally operate by exploiting relationships between neighbouring voxels, which are not present in point cloud data. 

The possibility of using deep networks to reason about points has been explored, resulting in networks that process point sets for the purpose of class labelling, part or semantic segmentation \cite{Qi2017,Qi2017a}. However, these networks either do not consider local information at all, instead relying on global pooling to embed relationships between individual points and the overall structure \cite{Qi2017}, or they extract local and global features by progressively growing scales in a hierarchical fashion \cite{Qi2017a}. The latter approach has proved to be effective for processing solid, predominantly convex shapes, such as animals and furniture, but is not optimal for thin, complex structures. This is because reasoning over local neighbourhoods of points may inadvertently include nearby, disconnected branches but not current branch end points. Additionally, we aim to extract meaningful topological information such as the location of junction points, the number of individual branches, and the branches to which each point belongs, which requires a loss function that is invariant to the number of branches. 

\begin{figure}[t]
	\centering
	\includegraphics[width=1\linewidth]{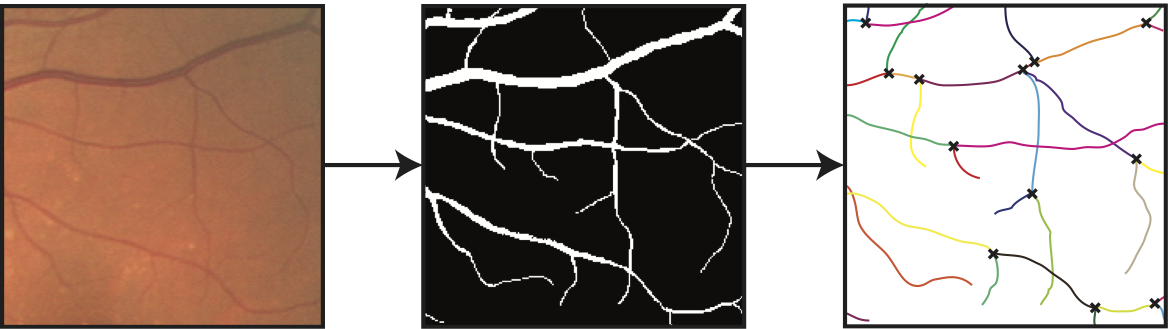}
	\caption{Workflow of instance segmentation of blood vessels in a retinal fundus image. After segmentation, centreline coordinates of vessels are extracted using skeletonisation, and input in random order to the proposed network. Right-most image shows the output of the network with point colour indicating branch membership, and identified branch end points indicated by crosses. (Best viewed in colour)}
	\label{fig:RetinaExample}
	\vspace{-1.5em}
\end{figure}

To meet these challenges, we present a data-driven topology estimation deep neural network that operates on unordered 2D or 3D centreline coordinates of thin, tubular, branching structures (Figure \ref{fig:RetinaExample}). We train our network using a novel multi-task learning framework to jointly develop branch and end-point identification abilities. Branches and end-points are identified using standard point-wise softmax loss, while individual branch instances are separated with a discriminative loss function \cite{DeBrabandere2017} operating at the point level. The latter allows the network to cluster an arbitrary number of branches, which can be identified in a post-processing step. The proposed network architecture, named BranchNet, is applied to the entire structure at once, helping it to develop a global reasoning and context for branch membership, but has local feature aggregation to inform junction localisation. Due to the intensive nature of the annotations required to train and validate the method, we train the network solely with synthetic data. However, we show that by applying domain randomization \cite{Tobin2017, Tremblay2018} in the data generation step, the performance translates well to real blood vessel structures. The key contributions of our work are as follows: 
\begin{itemize}[noitemsep,nolistsep]
	\item We propose a simple but effective means of embedding global structural information in point-based neural networks.
	\item We introduce a novel framework which enables the application of instance segmentation, the task of simultaneously solving object detection and semantic segmentation, to unordered point-cloud data. To the best of our knowledge, there are no prior works focusing on this problem.
	\item We employ extensive domain randomization during point-cloud generation to successfully transfer the networks success on to real data, without domain adaptation, fine-tuning, or having been exposed to real data during training.
	\item We provide a thorough sensitivity analysis showing that our proposed method is robust to multiple types of noise, with equal performance on 2D and 3D structures.
	\item We conduct an ablation study to investigate the importance of each domain randomisation parameter on the transfer to real data.
\end{itemize}

\section{Related Work}
\label{sec:related_work}
\subsection{Unstructured Point Clouds}
Point sets are unordered and not structured in a grid, therefore do not inherently favour a deep learning-based approach to processing. Therefore, for analysis with CNN's, previous works have transformed point clouds into uniform grids through rasterization \cite{Wu2015, Wu2016}, or represented them in kd-tree format \cite{Klokov}. More recent work has made progress towards directly consuming point-sets within a deep-learning framework. A pioneer in the field was PointNet, which returned a whole-set class label, or point-wise segmentation labels for a given point set \cite{Qi2017}. However, the particularly inspiring work was PointNet++, which introduced a hierarchical feature learning architecture to mimic the progressively growing receptive fields of convolutional neural networks, enabling the network to capture local geometry context. We further improve this network structure by including global feature descriptors along with local, and defining a novel multi-task loss function to enable instance segmentation.

\subsection{Instance Segmentation}
A key difficulty associated with the naive application of a softmax cross-entropy loss function to branch instance identification is the requirement for a constant number of branches (and therefore classes), with a specific hierarchical structure, which may not hold true for particular structures, such as trees in nature. This could be addressed by imposing an upper limit to the number of branches that can be detected, however this would limit the representational power of the network, and introduce imbalances in class representation. Therefore, we propose the use of a discriminative loss function to enable semantic instance segmentation of points into branches. Our loss function is inspired by \cite{DeBrabandere2017}, where pixels in a masked image were mapped to a location in feature space close to other pixels composing the same object, but away from those representing others. Separate instances are then identified with a post-processing operation. Other forms of instance segmentation have also been proposed, including pipelines with region or object proposals followed by segmentation \cite{Dai2016,Romera-Paredes2016}, and end-to-end recurrent neural networks that perform object detection and segmentation \cite{Stewart2016, Park2016, Romera-Paredes2016}. In contrast to these works, our method treats the input structure holistically, which is required in case of crossing branches, and is conceptually simpler and easier to implement than recurrent networks.

\subsection{Extraction of Topological Information}
Several methods exist for extracting meaningful topological information from thin, tree-like structures, however no generic frameworks perform equally on both 2D and 3D structures. Successful approaches for 2D data tend to include multi-stage pipelines where regions of interest are identified and then classified. These include \cite{Nougrara2016} and \cite{Abbasi-Sureshjani2016}, where directional filters to identify regions of interest are followed by optimisation to classify, and \cite{Pratt2017}, where a convolutional neural network is used to identify patches of interest, followed by a further neural network to classify said patches. By taking as input larger section of structures our network is able to develop a global reasoning and context for bifurcation locations and branches.

Previous approaches applied to 3D data have involved combined segmentation and topology estimation by growing an area outwards from a seed point, guided by image-derived constraints \cite{Choromanska2012,Gala2014, Sui2014}. This approach can experience early termination in the presence of intensity inhomogeneity, and noisy images. In contrast, our algorithm does not operate on seed points, and is therefore not negatively influenced by suboptimal seed point placement. Additionally, since our network is not informed by image-based features, it is not impacted by circumstances where local image features indicate a discontinuity, or the radius experiences local changes. Since our model operates on only a skeletonised branched structure, it is task-independent, and could in theory be applied as a post-processing step following segmentation and skeletisation, in any domain involving branching structures. Furthermore, our network is a generic framework that can operate on both 2D and 3D structures. 

\subsection{Training on Synthetic Data}
Use of synthetic data to train neural networks is an attractive proposition, particularly for work requiring complex annotations, where large, fully-annotated datasets are not available. Previous image-based work has attempted to make the generated data match the real data as closely as possible, by using high quality renderers \cite{Johnson-Roberson2017}. However, this often still fails to accurately match the statistics of real data. Several approaches have emerged to tackle the problem of domain adaptation, including re-training in the target domain \cite{Yosinski2014}, learning invariant features between domains \cite{Tzeng2014}, and learning a mapping between domains \cite{Taigman2016}. Alternatively, domain randomization has been proposed \cite{Tobin2017}, where synthetic data is generated with sufficient variation at training time that the network is able to generalize to real-world data at test time. This process removes the need for domain adaptation, and has been used successfully for object detection and classification \cite{Peng2015a, Tobin2017, Tremblay2018}, and training robotic control processes \cite{Tobin2017a, Peng2017}. This approach is particularly well suited to our application, since large, well-annotated datasets are not available, and randomization can be easily introduced in to the generation process of branching structures through structural changes, point jittering and dropout.

\section{Proposed Method}
\label{sec:method}
Suppose a thin, branching structure is represented by an unordered set of 3D points $\{P_i|\in \mathbb{R}|i=1,2,...,N\}$, where each point $P_i$ is a vector of its $(x,y,z)$ 3D location along the centreline of the structure. We are interested in the topology of the structure represented by these points, including which points belong to each branch, and the location of branch end points, in the form of semantic labels. Towards this end, we propose BranchNet, a deep neural network that directly consumes unordered points along the centreline of a branching structure, to identify the topology of the represented structure in a single-shot. 
\subsection{Network Architecture}
\begin{figure}[t]
	\includegraphics[width=1\linewidth]{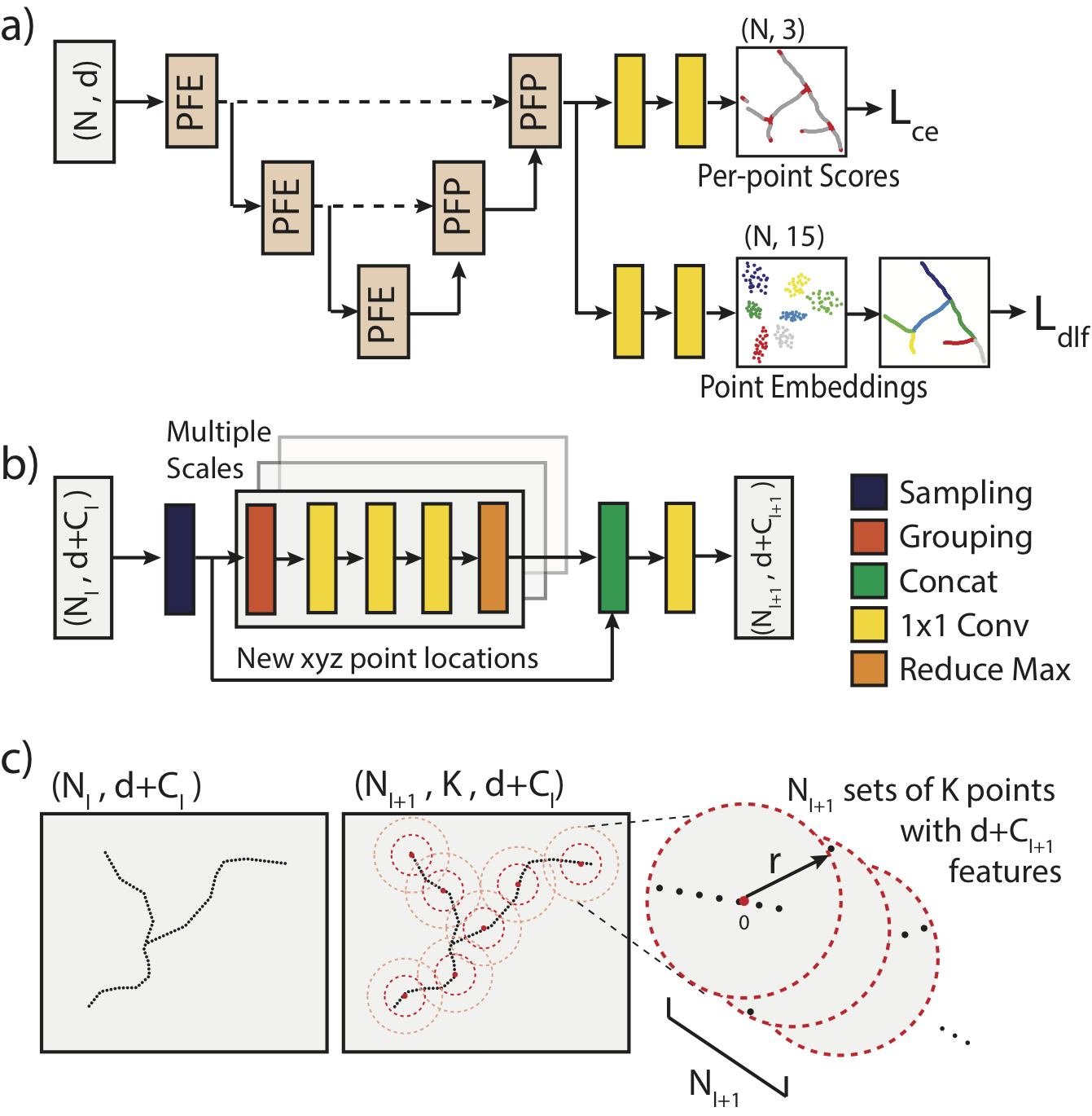}
	\caption{Illustration of the proposed BranchNet architecture using a 2D tree-like structure as an example. a) The overall hierarchical architecture with task specific shared fully connected layers and objectives. Dashed lines indicate skip connections. b) Expanded view of the Point Feature Embedding (PFE) layer. c) Graphical representation of sampling procedure. (Best viewed in colour)}
	\label{fig:network}
	\vspace{-1em}
\end{figure}

The structure of the network is shown in Figure \ref{fig:network}a.  Our network is composed of two main components: \emph{Point Feature Embedding (PFE)} and \emph{Point Feature Propagation (PFP)}, which can be considered as an encoder and a decoder, respectively, and are further explained in the following.  
\subsubsection{Point Feature Embedding}
The PFE component (Figure \ref{fig:network}b) maps the raw 3D coordinates to a feature space by hierarchically aggregating information at multiple scales. It is inspired by the set abstraction process described in \cite{Qi2017a}, but includes global feature descriptors along with local. The input to a PFE component $l$ is a matrix of size $N_l \times (d+C_l)$, where $N_l$ is the number of points with $d$-dim coordinate and $C_l$-dim point features. It outputs a $N_{l+1}\times (d+C_{l+1})$ matrix of $N_{l+1}$ subsampled points, selected using farthest point sampling from the original set, with $C_{l+1}$ new point features summarising local and global context (Figure \ref{fig:network}c). The local neighbourhood around each point $N_{l+1}$ is encoded by translating the $K$ points in a radius of $r$ around each point into a frame relative to the centroid point. The feature vector of each point is updated with three shared fully connected (FC) layers, similar to $1\times1$ convolutions, followed by max pooling over the $K$ points. We use Multi-Scale Grouping proposed in \cite{Qi2017a} to acquire point features at multiple scales. However, in contrast to \cite{Qi2017a}, we then concatenate the local features at each of the scales with the Cartesian coordinates of each sampled point, followed by a further shared FC layer. These locations are used as global point features, and serve to inform the network of relative point positions. Additionally, this small skip-connection aids in the training by improving gradient flow. 

\subsubsection{Point Feature Propagation}
To obtain features for all the original points, we propagate features from the subsampled points to the original points, as described in \cite{Qi2017a}. Briefly, point features are interpolated from $N_{l}$ points to $N_{l-1}$ points, and then concatenated with the skip-link point features, see \cite{Qi2017a} for more details. The features for each point are then updated using two (for the first two PFP layers) or three (for the final PFP layer) shared FC layers.

\subsubsection{Task-dependent Layers}
Feature propagation back to the original points is followed by several task-specific layers, without shared weights, as shown by yellow blocks in Figure \ref{fig:network}a). BranchNet is tasked with grouping points based on which branch they lie within a structure, as well as which points are close to branch end points. These tasks are separate but complimentary. Intuitively, the most difficult points to group are those near junctions (where multiple branches meet). Tasking part of the network with specifically locating these points forces attenuation, resulting in better branch cluster separation.

\subsection{Multi-Task Learning}
The BranchNet loss function comprises two parts: loss of point-wise semantic segmentation of branch end points from branches ($L_{ce}$), and the loss associated with instance segmentation of branches ($L_{dlf}$),
\setlength{\belowdisplayskip}{3pt} \setlength{\belowdisplayshortskip}{4pt}
\setlength{\abovedisplayskip}{3pt} \setlength{\abovedisplayshortskip}{4pt}
\begin{align}
L= L_{ce}+wL_{dlf}\; ,
\label{eq:eq1}
\end{align}
where $w$ is a weighting parameter to control the trade-off between the two components
of the loss.
During training, we pad our point sets such that the training set has a consistent number of points $(10,000)$. Therefore, our dataset contains three classes of points: padding, branch and end-point. $L_{ce}$ is calculated using point-wise cross-entropy on the soft-max output of the upper branch in Figure \ref{fig:network}a).

We use a discriminative loss function \cite{DeBrabandere2017} to calculate the instance segmentation loss $L_{dlf}$ (Eq. \ref{eq:dlf1}). This method allows for single-shot instance detection with feed-forward networks, avoiding the inefficiencies of detect-and-segment approaches. This loss function encourages points with the same label (therefore the same instance) to be projected to nearby locations in feature space, while embedding's with different labels would be represented far apart. This is achieved by applying two competing forces on an embedding: a variance force ($L_{var}$, Eq. \ref{eq:dlf2}) that pulls points towards their cluster mean, and distance force ($L_{dist}$, Eq. \ref{eq:dlf3}) pushing different clusters apart. Here, $C$ is the number of clusters, $\mu_c$ is the mean of cluster $c$ which has $N_c$ elements, and $x_i$ is an embedding. These forces are hinged (denoted by $[x]_+ = \mathrm{max}(0, x)$) such that embedding's within a distance of $\delta_v$ from their cluster centre are not pulled towards it, and those further than $\delta_d$ away are not repulsed. The overall loss, also contains a regularising term that pulls all clusters towards the centre.
\begin{align}
L_{dlf}=\alpha \cdot L_{var}+\beta \cdot L_{dist}+\gamma \cdot \frac{1}{C}\sum_{c=1}^{C}\lVert \mu_c \lVert~,
\label{eq:dlf1}
\end{align}
\begin{align}
L_{var}=\frac{1}{C}\sum_{c=1}^{C}\frac{1}{N_c}\sum_{i=1}^{N_c}[\lVert \mu_c - x_i \lVert - \delta_v]^2_+~,
\label{eq:dlf2}
\end{align}
\begin{equation}
\begin{split}
L_{dist}=\frac{1}{C(C-1)}\sum_{c_A=1}^{C}\sum_{c_B=1}^{C}[2\delta_d-\lVert \mu_c - x_i \lVert]^2_+ \qquad \\
\forall \quad c_A \neq c_B~.
\label{eq:dlf3}
\end{split}
\end{equation}

This formulation results in all point embedding's being located within $\delta_v$ from the centre of their associated cluster, and at least $2\delta_d$ from the centres of all other clusters, provided the loss has converged. At test time this assumption may not be accurate, so we apply a fast variant of the mean-shift algorithm \cite{Fukunaga1975} to locate and threshold around cluster centres.

We train the network with margins of $\delta_v=0.7$, and $\delta_v=1.5$, and 15 output dimensions. The weight parameters were set to $w=0.05$ (for Eq. \ref{eq:eq1}), $\alpha=1.5$, $\beta=1$ and $\gamma=0.001$ (for Eq. \ref{eq:dlf3}). The Adam training algorithm \cite{Kingma2014a} was used for end-to-end training with mini-batch size of 12 and learning rate of 1e-5.

We use PFE, PFP and TD to represent point feature embedding, point feature propagation, and task dependent layers, respectively. Parameters for each of the layers in BranchNet are shown in Table \ref{tab:Parameters}, with $m$ scales, $K$ local regions of ball radius $r$. $[l_1,\ldots, l_d]$ represents the $d$ FC layers with width $l_i$ $(i = 1, \ldots,d)$, and $g$ is the width of the FC layer following concatenation of local point features with global location. Both TD branches acquire features from the final PFP layer, and $R=4e-3$. All FC layers are followed by batch normalization and ReLU except for the last layer in each TD path.
\begin{table}[h]
	\centering
	\caption{BranchNet parameters}
	\label{tab:Parameters}
	\small\addtolength{\tabcolsep}{1pt}
	\begin{tabular}{@{}lcccccc@{}}
		\toprule
		\# & Layer Type & K   & m & r   & $l^1$,...,$l^d$   & g  \\
		\midrule
		\multirow{3}{*}{1}  & \multirow{3}{*}{PFE}  & \multirow{3}{*}{512}   & 1 & 1R  & 32, 32, 64  & \multirow{3}{*}{128}  \\
		&   &  & 2 & 4R  & 64, 64, 128   &  \\
		&  &  & 3 & 16R & 64, 96, 128   & \\ \midrule
		\multirow{3}{*}{2} & \multirow{3}{*}{PFE}  & \multirow{3}{*}{128}   & 1 & 4R  & 64, 64, 128  & \multirow{3}{*}{256}  \\
		&   &  & 2 & 16R  & 128, 128, 256   &  \\
		&   &  & 3 & 64R &  128, 128, 256    & \\ \midrule
		3 &PFE  & 1		  & - & -  & 256, 512, 1024  & - \\	\midrule
		4 &PFP  & -		  & - & -  & 256, 128  & - \\	\midrule
		5 &PFP  & -		  & - & -  & 128, 128, 128  & - \\	\midrule
		6 &TD1  & -		  & - & -  & 128, 15 & - \\	\midrule
		7 &TD2 & -		  & - & -  & 128, 3  & - \\	
		\bottomrule
	\end{tabular}
\vspace{-1.5em}
\end{table}

\section{Experiments}
\label{sec:experiments}
We train our model on synthetically generated, branching tree structures (Figures \ref{fig:JitterAndDropoutFigure} and \ref{fig:examples}). Both 2D and 3D structures are used to train separate networks, which are evaluated using real data.

\subsection{Synthetic Data}
\begin{figure}[t]
	\centering
	\includegraphics[width=1\linewidth]{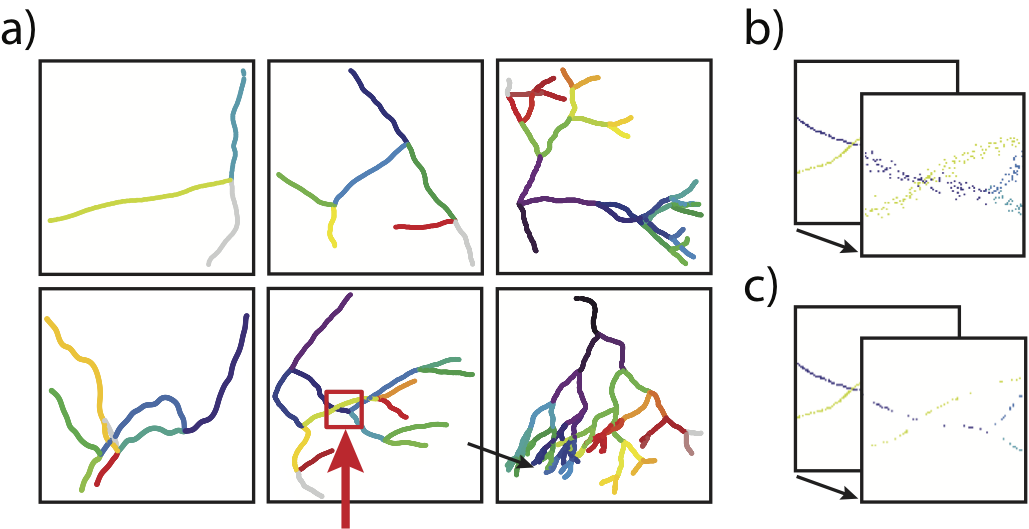}
	\caption{Examples of branching structures used as inputs to the network. Points are coloured according to their branch membership. b,c) Zoomed in versions of the region indicated by the arrow, with varying amounts of point jitter (b, jitter standard deviations shown here are 0-4 pixels) and point dropout (c, 0-75\% dropout shown). (Best viewed in colour)}
	\label{fig:JitterAndDropoutFigure}
	\vspace{-1.5em}
\end{figure}

 Synthetic data provides the opportunity to train our approach on highly complex structures similar to those existing in nature, such as neurons and blood vessels, but without undergoing the extensive manual annotations required herein. Domain randomization during data generation results in a large range of inputs, compared to which real data is considered just another variation. Training data was generated by a recursive branching process, with random step lengths and angles to simulate structures with varying degrees of tortuosity. Each structure begins with the same seed point ($x=(0,0,0)$) and growth direction ($\theta=0$, $\phi=0$ using spherical coordinates). Trees were then grown by randomly varying the following aspects of morphology:
\begin{itemize}[noitemsep,nolistsep]
	\item Number of branching levels (up to 4 levels, chosen with equal probability)
	\item Bifurcation or trifurctation (80\% versus 20\% of junctions respectively)
	\item Start angle of each branch (uniform probability from  $[\SI{0}{\degree}, \SI{90}{\degree}])$
	\item Number of steps in each branch (Gaussian distribution, mean of 20, standard deviation of 8)
	\item Angle of each step (Gaussian distribution with mean equal to the angle of the previous step and standard deviation drawn with uniform probability from $[\SI{10}{\degree}, \SI{60}{\degree}]$)
	\item Length of each step (uniform probability from $(0,1]$)
\end{itemize}
Randomisation of both the number of steps in a branch and the length of each step produces branches with varying degrees of tortuosity. A spline is then fitted through the points, with clamped end points and angles. Example structures are shown in Figure \ref{fig:JitterAndDropoutFigure}a. For 2D data, only $\phi$ is changed each step with $\theta=0$.
The generated structures were randomly rotated and scaled to lie within a $512\times512\times512$ cube (or square for 2D data), and rounded to the nearest integer (to approximate voxels or pixels). Duplicate points, such as those occurring at cross-over points, were removed, and the remaining points jittered. For training, we applied point jittering by randomly sampling point offsets from a Gaussian with standard deviation of 3 pixels, and point dropout with a probability of $0.4$. The effect of varying levels of point jitter and dropout (examples shown in Figures \ref{fig:JitterAndDropoutFigure}b and c) were tested during evaluation on 1000 unseen randomly generated structures. Point sets were then scaled to unit width, and point order randomised. This produces an infinite data source of thin tree-like structures that are unorganized and scattered, with unknown connectivity.

\subsection{Real Data} %
We tested the trained 2D and 3D networks on two separate, publicly available datasets. The 2D network was evaluated on retinal vasculature segmented from fundus images in the DRIVE dataset \cite{Staal2004}. For 5 images within the test dataset, branches and branch end points were annotated manually by a human observer and checked for accuracy by a trained expert. Rather than using ground truth segmentations as input, we use the output of a pre-trained U-net\footnote{https://github.com/orobix/retina-unet} so as to encompass any inaccuracies arising from segmentation (such as disconnected branches), thereby mimicking the use of BranchNet in a real use case. The centreline was then obtained using skeletonisation. This process is shown visually in Figure \ref{fig:RetinaExample}. 36 overlapping patches were taken from each of the images to limit the number vessels shown within each patch, and to increase the size of the test set. 
The 3D network was evaluated on a centreline model of left and right coronary trees from a segmentation of a computed tomography scan (CTCA) \cite{vasmodel}. The left and right trees are treated as separate point sets, resulting in a dataset size for testing of 2. Patches were not obtained to extend the dataset (as per the 2D data), due to the limited number of branches available in each set. Therefore, to artificially enhance the size of this dataset, the prediction for each structure was repeated 50 times, with randomised rotations of the structure.
Point sets for both datasets were then scaled to unit width, and point order randomised.

\subsection{Evaluation Metrics}
To evaluate the success of instance segmentation we report two metrics defined in \cite{Scharr2016}: Symmetric Best Dice (SBD) and Absolute Difference in Count (DiC).
Symmetric Best Dice is computed by first finding, for each input label, the ground truth label yielding the maximum Dice score. These scores are then averaged over input labels, and termed the ``Best Dice" score. The ``symmetric" element is introduced by repeating the process for each ground truth label paired with its most favourable input label, the minimum Best Dice score between the measures is the SBD. The Absolute Difference in Count is the absolute mean of the difference between the predicted number of branches and the ground truth over all images (lower is better). 
In a test situation, the location of each individual bifurcation may be required, which is not immediately accessible if multiple points describe each bifurcation. Therefore, we identify clusters of points labelled at bifurcations by performing mean-shift clustering on these points. Ideally, cluster centres will align with actual bifurcation locations. To quantify the success of bifurcation localisation we define the Cluster-based Dice Score ($\mathrm{DS_C}$):
\begin{align}
DS_C=\frac{2~{TP}_C}{2~{TP}_C + {FP}_C + {FN}_C},
\end{align}
where ${TP}_C$, ${FP}_C$ and ${FN}_C$ respectively indicate the number of ground truth clusters correctly predicted, clusters predicted incorrectly, and ground truth clusters missed in the prediction.

\section{Results and Discussion}
\label{sec:results}

\subsection{Comparison of Architectures}
In this section we compare the performance of the proposed BranchNet (BNet) to those of PointNet++ with Multiscale Grouping (PN++) \cite{Qi2017a}, appended with the proposed task-dependent layers and multi-task loss function. PointNet Vanilla \cite{Qi2017} and PointNet++ with Multiresolution Grouping \cite{Qi2017a} were also tested, but showed reduced performance for all tasks, and are therefore not shown here for brevity. 

\subsubsection{Synthetic Data}
\begin{figure}[t]
	\centering
	\includegraphics[width=1\linewidth]{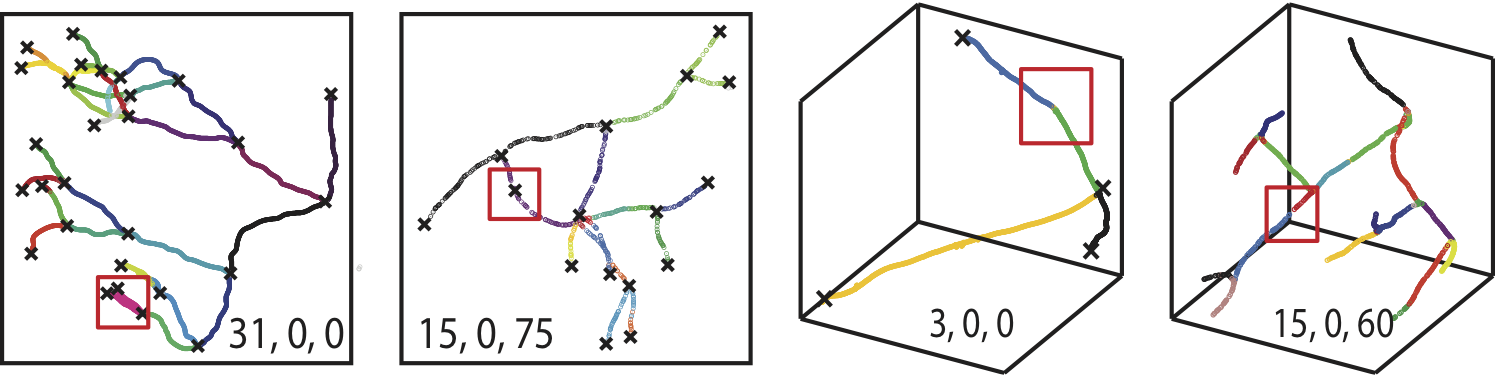}
	\caption{Examples of BranchNet outputs. Points are coloured according to their predicted branch membership, crosses indicate predicted bifurcation and end point locations. Numbers shown in each image indicate: Number of branches, jitter standard deviation (in pixels), point dropout percentage. Red boxes highlight mistakes of interest. (Best viewed in colour)}
	\label{fig:examples}
	\vspace{-1.5em}
\end{figure}

We investigate the impact of varying structural complexities, noise and sampling density, by increasing the numbers of branches in the synthetically generated structures, and amounts of point jitter and point dropout, respectively. Example results for our method applied to synthetic data during test time are shown in Figure \ref{fig:examples}. 

To investigate the robustness of the network to structural complexity, we tested it against 1000 structures at each of 4 different levels of branching (bifurcations only), but no jittering or dropout. Results are shown in Table \ref{tab:complexity}. 
Interestingly, SBD results show that instance segmentation was more accurate for mid-range complexity structures. This is likely due to the longer branch length of simpler structures, resulting from structure normalisation prior to the network input. This may cause the entire branch to not be covered by the lowest resolution of grouping, and be falsely considered as multiple branches, as shown in Figure \ref{fig:examples}. However, bifurcations are clearly more difficult to accurately identify with higher numbers of branches, as is evidenced by the lower $\mathrm{DS_C}$ results for higher complexity levels. This is unsurprising, given that the likelihood of crossover points (which are a confounding factor) increases with branch numbers. The results also show that for all complexity levels our network outperforms PointNet++.

\begin{table}[h]
	\centering
	\caption{Structural Complexity Results}
	\label{tab:complexity}
	\small\addtolength{\tabcolsep}{1pt}
	\begin{tabular}{@{}ccccccccccccccc@{}}
		\toprule
		& & \multicolumn{2}{c}{2D} &   \multicolumn{2}{c}{3D} \\  
		\cmidrule(l){3-4} \cmidrule(l){5-6}
		&\#  Branches & PN++ & BNet  & PN++ &  BNet \\  \midrule
		\multirow{4}{*}{SBD}  	
											& 3 & 52.4 & \textbf{55.2} & 30.0 & \textbf{48.0} \\
											& 7 & 52.5 & \textbf{57.5} & 42.0 & \textbf{71.1} \\
											& 15 & 54.5 & \textbf{67.7} & 46.6 & \textbf{61.3} \\
											& 31 & 54.2 & \textbf{67.5} & 45.1 & \textbf{50.1} \\ \midrule
		\multirow{4}{*}{DiC}  	
											& 3 & 3.2 & \textbf{1.2} & 11.0&  \textbf{1.4}\\
											& 7 & 7.5 & \textbf{2.9} & 13.4& \textbf{3.2}\\
											& 15 & 9.6 & \textbf{3.4} & 15.7 &  \textbf{4.4}\\
											& 31 & 10.0 & \textbf{7.6} & 14.5 & \textbf{12.1}\\ \midrule
		\multirow{4}{*}{$\mathrm{DS_C}$}  	
											& 3 & 78.0 & \textbf{84.1} & 98.7 &  \textbf{99.0}\\
											& 7 & 97.3 & \textbf{99.8} & 98.6 & \textbf{98.9}\\
											& 15 & 95.6 & \textbf{98.9} & 98.3 &  \textbf{98.8}\\
											& 31 & 89.8 & \textbf{94.6} &  97.0 & \textbf{97.9}\\ 
	\bottomrule
	\end{tabular}
\vspace{1em}
\caption{Sampling Density Results, $\# \mathrm{Branches}=15$}
\label{tab:density}
	\begin{tabular}{@{}ccccccccccccccc@{}}
		\toprule
		& & \multicolumn{2}{c}{2D} &   \multicolumn{2}{c}{3D} \\  
		\cmidrule(l){3-4} \cmidrule(l){5-6}
		&Dropout (\%) & PN++ & BNet  & PN++ &  BNet \\  \midrule
		\multirow{4}{*}{SBD}  	
											& 30 & \textbf{67.8} & 67.6 & 41.3 & \textbf{63.0} \\
											& 45 & \textbf{67.5} & \textbf{67.5} & 40.8 & \textbf{62.7} \\
											& 60 & 66.3 & \textbf{68.1} & 39.7 & \textbf{61.2}\\
											& 75 & 62.5 & \textbf{67.8} & 37.1 & \textbf{56.1}\\ \midrule
		\multirow{4}{*}{DiC}  	
											& 30 & 4.0 & \textbf{2.9} & 4.3 & \textbf{2.7} \\
											& 45 & 4.3 & \textbf{3.0} & 6.7 & \textbf{2.9} \\
											& 60 & 4.4 & \textbf{2.9} & 8.1 & \textbf{2.6} \\
											& 75 & 5.1 & \textbf{3.4} & 11.2 & \textbf{3.2} \\ \midrule
		\multirow{4}{*}{$\mathrm{DS_C}$}  	
											& 30 & 98.7 & \textbf{99.0} & 98.6 & \textbf{98.9} \\
											& 45 & 98.6 & \textbf{98.9} & 98.3 & \textbf{98.8} \\
											& 60 & 98.3 & \textbf{98.8} & 97.0 & \textbf{97.9} \\
											& 75 & 97.0 & \textbf{97.9} & 86.5 & \textbf{87.6} \\ 
		\bottomrule
	\end{tabular}
\vspace{1em}

\caption{Sampling Noise Results, $\# \mathrm{Branches}=15$}
\label{tab:noise}
	\begin{tabular}{@{}ccccccccccccccc@{}}
		\toprule
		& & \multicolumn{2}{c}{2D} &   \multicolumn{2}{c}{3D} \\  
		\cmidrule(l){3-4} \cmidrule(l){5-6}
		&St. Dev (px) & PN++ & BNet  & PN++ &  BNet \\  \midrule
		\multirow{4}{*}{SBD}  	
											& 1 & 50.6 & \textbf{67.2} & 35.4 & \textbf{62.5}  \\
											& 2 & 49.7 & \textbf{68.4} & 34.1 & \textbf{61.7} \\
											& 3 & 46.9& \textbf{72.5} & 32.7 & \textbf{64.6} \\
											& 4 & 41.5 & \textbf{78.6} & 30.0 & \textbf{65.5} \\ \midrule
		\multirow{4}{*}{DiC}  	
											& 1 & 7.0 & \textbf{3.7} & 5.9 & \textbf{4.7}\\
											& 2 & 13.9 & \textbf{3.4} & 11.1 & \textbf{5.5}\\
											& 3 & 16.8 & \textbf{2.9} & 20.4 & \textbf{6.8}\\
											& 4 & 22.3 & \textbf{7.2} & 29.0 & \textbf{13.7}\\ \midrule
		\multirow{4}{*}{$\mathrm{DS_C}$}  	
											& 1 & 94.6 & \textbf{99.2} & 76.6 & \textbf{78.0}\\
											& 2 & 94.2& \textbf{99.4} & \textbf{97.3} & \textbf{97.3}\\
											& 3 & 90.5 & \textbf{98.7} & 95.1 & \textbf{95.7}\\
											& 4 & 77.8 & \textbf{92.7} & 68.4 & \textbf{89.8}\\ 
		\bottomrule
	\end{tabular}
	\vspace{-1.5em}
\end{table}

 We randomly drop increasing numbers of points during test time to validate our network's robustness to non-uniform and sparse data (results shown in Table \ref{tab:density}). For this section and the next, we tested against 1000 structures (at each perturbation level) with the 3rd level of branching (bifurcations only, 15 branches in total). Our network outperformed PointNet++, however the results were considerably closer. This is unsurprising since both models select point neighbourhoods at multiple scales, and learn how to weight them. This means that even if nearby points are missing, the available points can agglomerate features with more distant points. These results are particularly impressive given that this level of point dropout would most likely severely disrupt any traditional means of branch tracing.

To evaluate our network's sensitivity to noisy data we jitter the point locations by sampling from zero-centred Gaussians with increasing standard deviations (results shown in Table \ref{tab:noise}). Jittering reduces the information inherent in local point features, which has a particularly negative effect on thin, tortuous structures. By including global information in BranchNet, we allow the network to reason over nearby groups of points. This is borne out by the improved bifurcation localisation results for our method. Additionally, including global information ensures that branches with similar structural features (such as angle and tortuosity) are less likely to be mistakenly clustered together. This is shown by the improved performance in the instance segmentation task. However, there remains room for improvement, especially when considering the poor DiC results for 3D structures with large amounts of jitter, likely due to the model predicting that connected branches were split.
 
\subsubsection{Real Data}
\begin{figure}[tbp]
	\centering
	\includegraphics[width=1\linewidth]{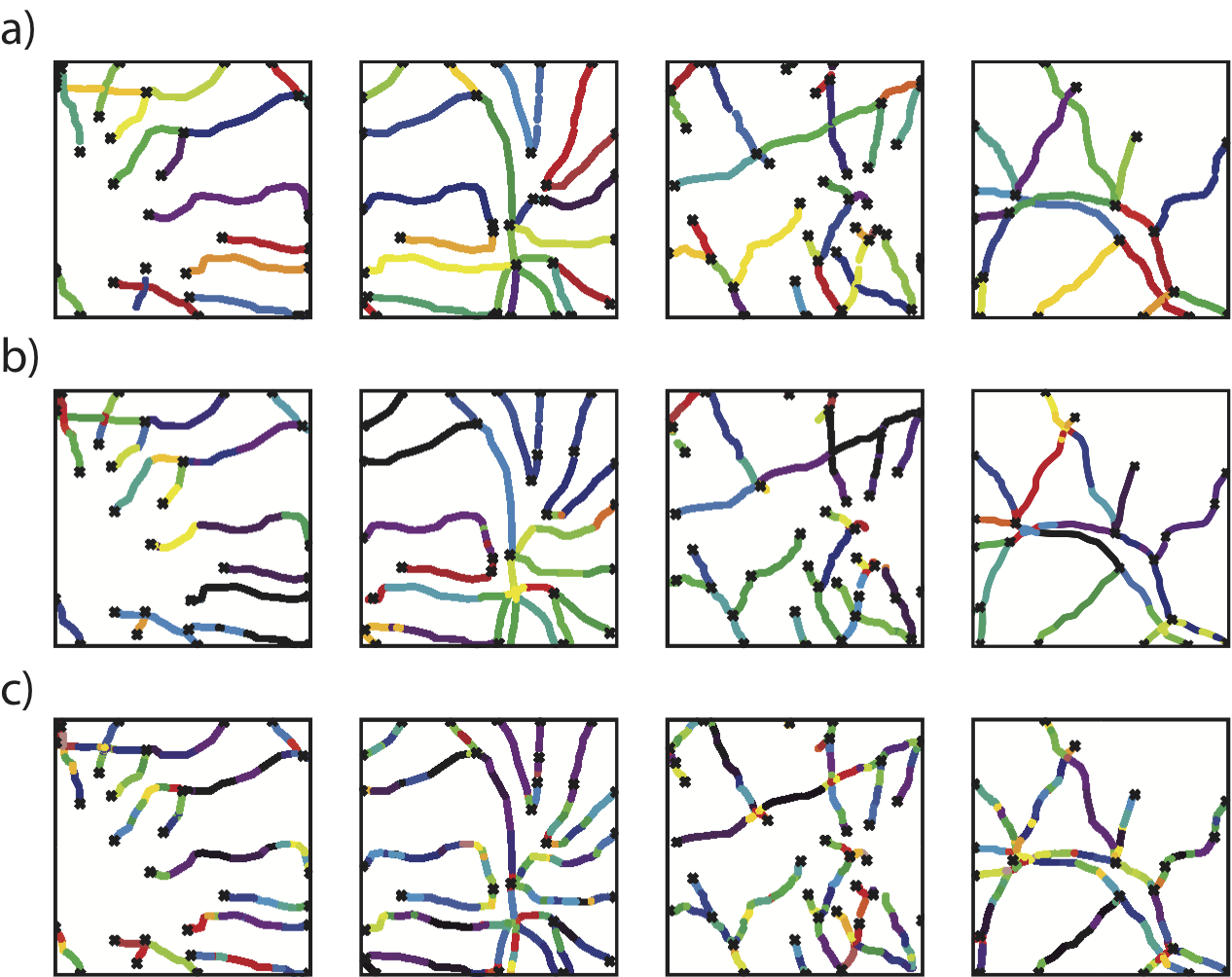}
	\caption{Examples from processing of real 2D retinal vasculature data. Displayed are the ground truth (a), the output of BranchNet (b) and the output of PointNet++ (c).}
	\label{fig:RealData_2D}
	\vspace{-1em}
\end{figure}

To evaluate the accuracy of our learned topology estimator in the real world, we use two datasets comprising branching networks of blood vessels. For the network trained for 2D data, we use patches of retinal vasculature obtained through segmentation of fundus images. Examples of the output of BranchNet and PointNet++ on this test set are shown in Figure \ref{fig:RealData_2D}b and c, respectively (ground truth shown in Figure \ref{fig:RealData_2D}a). Summary metrics are shown in Table \ref{tab:RealData}.

The results show that even though our model has never been exposed to real data, it is able to successfully identify individual branches and their end points, which is reflected in the SBD,DiC and DSC scores, which are on par with those for synthetic data. Furthermore, the results show that the model can distinguish between crossover points and bifurcations, and tends not to classify points near cross-overs as junctions. Impressively, the model is able to generalize to branches that are disconnected from the main structure (given that such branches were not included in the training set). Additionally, BranchNet visually outperforms PointNet++ (Figure \ref{fig:RealData_2D}c) on the basis of instance segmentation, as shown by the increased number of incorrectly identified branches in PointNet++. However, both networks produced comparable results for identification of junction points, though BranchNet produced less false positives.

\begin{table}[tbp]
	\centering
	\caption{Aggregated results on real data for PointNet++ (PN++) and BranchNet (BNet).}
	\label{tab:RealData}
	\small\addtolength{\tabcolsep}{1pt}
	\begin{tabular}{@{}lccccccccccccc@{}}
		\toprule
		& \multicolumn{2}{c}{2D} &   \multicolumn{2}{c}{3D} \\  
		\cmidrule(l){2-3} \cmidrule(l){4-5}
		& PN++ & BNet  & PN++ &  BNet \\  \midrule
		SBD & 34.6 & \textbf{59.3} & 44.4 & \textbf{63.7} \\
		DiC & 38.6 & \textbf{4.14} & 6.09 & \textbf{4.23} \\
		DSC & 85.6& \textbf{94.3} & 82.2 & \textbf{95.5}  \\\bottomrule
	\end{tabular}
	\vspace{-1.5em}
\end{table}

\begin{figure}[tbp]
	\centering
	\includegraphics[width=1\linewidth]{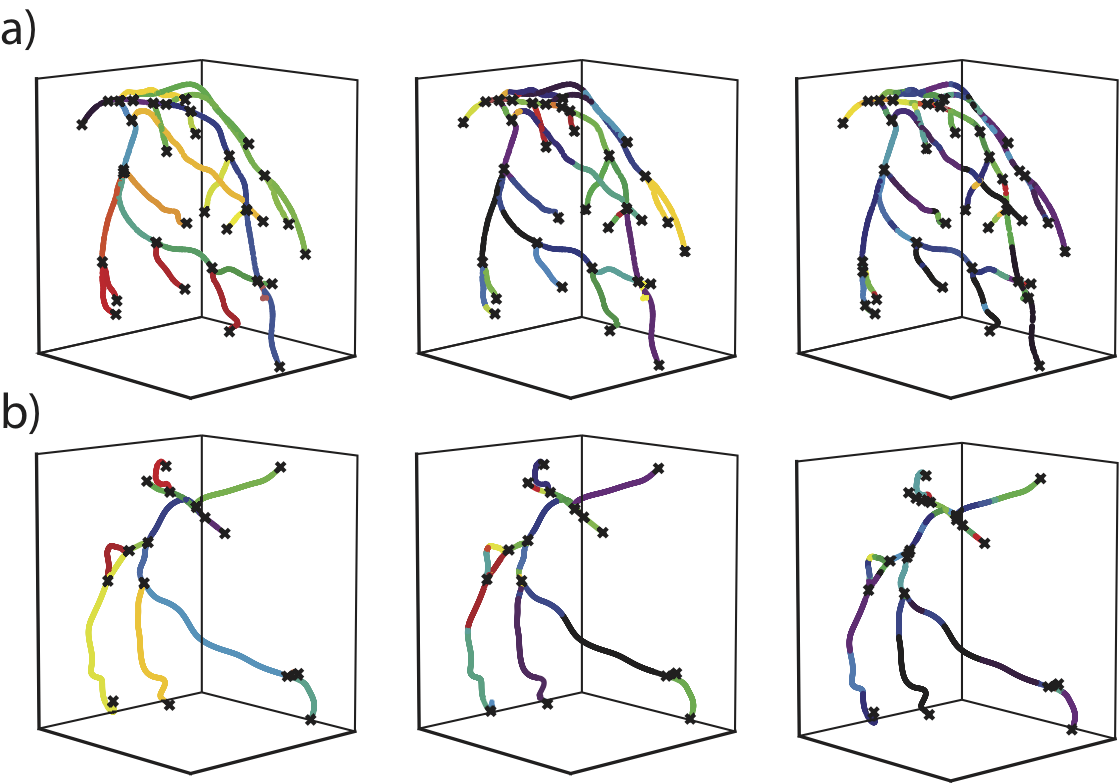}
	\caption{Examples from processing of real 3D coronary artery data. Displayed are the ground truth (column 1), the output of BranchNet (column 2) and the output of PointNet++ (column 3), for both structures in the dataset (a and b).}
	\label{fig:RealData_3D}
	\vspace{-1em}
\end{figure}

Examples of the performance of BranchNet on real 3D data are shown in Figure \ref{fig:RealData_3D}. The network successfully identified all branch end points. However, sometimes it also classified points mid-way along branches incorrectly, resulting in bifurcation false negatives, and lowering the Cluster-based Dice Score. This was more common for PointNet++.

The most common failure of the model is splitting a single branch in to multiple instances. This failure was also apparent on synthetic data, and may be due to the lowest resolution of point grouping in the network not covering the entire branch. This particular failure points to an inherent limitation of single shot neural networks in this application. Future work will investigate methods of solving this by imbuing the network with more power to reason over different scales, such as the multi-branch convolutional network used by \cite{Yu2018} for the purpose of super-resolution. Additionally, future studies will investigate the use of including other features in to the point representation to strengthen relationships between points on the same branch. For example, including further dimensions for radius (at each point along the centreline) and image-based features such as RGB values of the respective pixels.

\subsection{Ablation Study}
To study the effects of individual domain randomization parameters, we performed an ablation study by systematically omitting them one at a time. We used the same parameters as in the previous sections, but trained four additional versions of BranchNet for each 2D and 3D structures, each time fixing one of the following parameters in the synthetic data: branch length, number of branches (bifurcations only, 15 branches in total), point jitter (set to zero), or point dropout (set to zero). Table \ref{tab:Ablation} shows the results of omitting each of these individual components. We found that the method is sensitive to all of the factors, however the importance of each factor is dependent on dimensionality of the data. The success on 2D data was most dependent on randomisation of the number of branches, which reflects the large variability in number of visible branches in real data. 


The 3D data accuracy was most negatively affected by fixing branch length in the synthetic training data. This reflects the large range in branch length of the structures. The 3D network displayed the best performance on SBD when number of branches was fixed, despite the test structures having more branches than the fixed number (15). Given the small size of the evaluation dataset for 3D data, we can not be confident in the generalizability of our ablation test results. In the future, it would be preferable to perform the analysis using a larger dataset. 
Overall, relevant domain randomization parameters were clearly highly dependent on the particular dataset, but in most cases full randomization was most successful. However, performance was reduced compared to synthetic data. This suggests that the training structures do not fully encompass the range of real data, even with domain randomization. 
In future studies we will investigate further randomization parameters, and the inclusion of other types of thin, tortuous network structures, not limited to those branching from a single seed point.

\begin{table}[t]
	\centering
	\caption{Real 2D and 3D data test results for BranchNet trained on synthetic data with individual domain randomization parameters omitted. Normal refers to full parameter randomization.}
	\label{tab:Ablation}
	\small\addtolength{\tabcolsep}{1pt}
	\begin{tabular}{@{}clcccccccccccc@{}}
		\toprule
		& \multicolumn{3}{c}{2D} &   \multicolumn{3}{c}{3D} \\  
		\cmidrule(l){2-4} \cmidrule(l){5-7}
	    Fixed Param.	& SBD & DiC & 	$\mathrm{DS_C}$ & SBD & DiC & 	$\mathrm{DS_C}$ \\  \midrule
		normal 				& \textbf{59.3}& \textbf{4.14} & \textbf{94.3} 							& 63.7 & \textbf{4.23} & \textbf{95.5}		\\
		length 				& 54.2 & 5.42 & 91.5							& 59.5 & 7.21 & 86.4	  \\  
		jittering 			 & 54.1 & 8.13 & 93.6 							  & 62.1& 4.48 & 94.7	  \\    
		\# branches 	& 49.4 & 15.4 & 87.1							& \textbf{65.8} & 5.09 & 93.2	\\  
		dropout			  & 54.2 & 5.75 & 88.2 							   & 61.0 & 6.77 & 88.1		\\   \bottomrule
	\end{tabular}
\vspace{-1.5em}
\end{table}

\section{Conclusion}
\label{sec:conclusions}
In this work, we proposed a novel deep learning based approach to automatically extract topological information from unsorted point clouds of thin tree-like 2D and 3D structures. To analyse structures with arbitrary numbers of branches we proposed a novel, multi-task loss function, incorporating cross-entropy and discriminative losses. We trained the network using extensive domain randomization applied to synthetically generated tree-like structures, and showed that it successfully transferred to real data. In the future, it would be interesting to apply this network to higher dimensional spaces where CNN architectures would be computationally unfeasible, such as time-varying 3D structures. 

{\small
\bibliographystyle{ieee}
\bibliography{WACV2019}
}

\end{document}